\pdfoutput=1
\documentclass[11pt]{article}

\usepackage[]{acl}

\usepackage{hyperref}

\usepackage{times}
\usepackage{latexsym}

\usepackage[T1]{fontenc}
\usepackage[utf8]{inputenc}

\usepackage{microtype}

\usepackage{inconsolata}
\usepackage{booktabs}
\usepackage{multirow}
\usepackage{adjustbox}
\usepackage{makecell}
\usepackage{enumitem}
\usepackage{authblk}
\usepackage{enumitem}
\usepackage{float}
\usepackage{svg}

\title{CLIX: Cross-Lingual Explanations of Idiomatic Expressions}

\author{\textbf{Aaron Gluck}$^{1}$ \quad \textbf{Katharina von der Wense}$^{1,2}$ \quad \textbf{Maria Leonor Pacheco}$^{1}$ \\
    { $^1$University of Colorado Boulder \quad $^2$Johannes Gutenberg University Mainz} \\
    {\texttt{\{aaron.gluck, katharina.kann, maria.pacheco\}@colorado.edu}}
}

\begin{document}
\maketitle
\begin{abstract}

Automated definition generation systems have been proposed to support vocabulary expansion for language learners. The main barrier to the success of these systems is that learners often struggle to understand definitions due to the presence of potentially unfamiliar words and grammar, particularly when non-standard language is involved. To address these challenges, we propose CLIX, the task of Cross-Lingual explanations of Idiomatic eXpressions. We explore the capabilities of current NLP models for this task and observe that, while it remains challenging, large language models show promise. Finally, we perform a detailed error analysis to highlight the key challenges that need to be addressed before we can reliably incorporate these systems into educational tools. 

\end{abstract}

\section{Introduction}

The use of technology has become an important part of the language learning process \citep{doi:10.1080/09588221.2012.700315, doi:10.1080/09588221.2020.1744666}. One of the main areas of interest for the proponents of technology-assisted language learning is vocabulary expansion, where recent studies have demonstrated a significant impact in student engagement and increased vocabulary knowledge \citep{Fisher2016TheUO,Guaqueta2018TheUO,doi:10.1080/19345747.2021.1917028}. To support the development of these technologies, considerable work has been devoted to the study of automated definition generation \citep{ni_learning_2017,gadetsky_conditional_2018,ishiwatari-etal-2019-learning,bevilacqua-etal-2020-generationary}. 

The effective use of definition generation systems in second language vocabulary expansion is complicated by phenomena such as the coexistence of many possible meanings for a particular word or phrase \citep{enayati_impact_2020,TylerEvans+2003+95+160}, and the potential presence of unfamiliar words and grammar in the generated definitions \citep{zhang-discussion,kong-etal-2022-multitasking}. While we make no assertion that any particular word in a definition will be unfamiliar to a language learner, the fact remains that there \textit{might} be one or more such words. However, providing the language learner with an explanation in their native language may help remove this concern in many cases. In addition, a lot of these systems ignore the complexities of non-standard language such as idiomatic expressions, which are generally composed of words that combine to represent a meaning that does not equal the sum of its parts \citep{Nunberg1994-NUNI,kovecses-idioms,adelnia2011translation}.

Motivated by these challenges, we propose the task of Cross-Lingual explanation of Idiomatic eXpressions (CLIX). Given a source language idiomatic expression, we aim to \textbf{generate a natural language explanation in a specified target language}. As opposed to \textit{definition}, we use the term \textit{explanation} to allow for more flexibility in tailoring explanations to certain situations as well as potential inclusions of context. Rather than expecting a static, one-to-one mapping from idiom to definition, we allow models to approximate a one-to-many function. For example, explanations can be considered correct if they include elaborations, examples of use, and etymological information. %In general, two people would not express ideas in the exact same way, so why should a model be required to explain idiomatic expressions in a completely fixed manner? 
The core concern of this research is the efficacy of the explanation and the final outcome for a potential student. 

We additionally hypothesize that obtaining explanations of idiomatic expressions in the learner's first language removes many of the barriers to understanding introduced by traditional definition generation systems. We choose to focus on idiomatic expressions as they are an important element of language learning that is particularly challenging for learners and automated systems alike. Consider the utterance, \textit{he and I don't see eye to eye on a variety of topics}. The idiomatic expression contained within this sentence is not composed of particularly challenging words, yet in this instance it denotes a negative stance relative to the referent's opinions and its meaning has nothing to do with the general definitions of the words \textit{see} and \textit{eye}. 
% Additionally, it is clear that although the entire expression might, the individual words composing this idiomatic expression do not generally appear in complementary distribution with the word \textit{agree}.
This example helps to display the fact that idiomatic expressions often are of unpredictable form, appear in limited numbers of syntactic frames, include various types of figuration, indicate an affective stance, and can describe situations of social interest \citep{Nunberg1994-NUNI}.

To explore the capabilities and limitations of modern NLP systems for generating cross-lingual explanations of idiomatic expressions, we construct two datasets of English idioms and their explanations in English, Spanish, and German: EPIE-ME, an open dataset composed of 628 unique idioms with multi-lingual explanations built from the EPIE corpus~\citep{saxena_epie_2020}, and Oxford-ME, composed of 6,218 unique idioms taken from the Oxford Dictionary of Idioms 4th Edition \citep{Oxford}. We explore performance of both pre-trained sequence-to-sequence models and Large Language Models (LLMs), and show that our task configuration is much more challenging than previous definition generation tasks. 
% Will rework the part below once we have all results in. 
%In addition to this, we confirm some known properties of idiomatic expressions. Specifically, by contrasting the two models used for our experiments, 
%Unsurprisingly, we obtain the best performance with the model that has been exposed an immense amount of data, likely including some or all of the expressions in our dataset. 
We further provide some discussion on why na\"ive translation approaches do not suffice for this task.

%To combat the issues outlined above, we propose the task of Cross-Lingual explanation of Idiomatic eXpressions (CLIX), for which we will demonstrate two simple approaches before offering ideas for further work. We explore performance of both sequence-to-sequence models and Large Language Models (LLMs). We show that our task configuration is much more challenging than those that are similar, and confirm some known properties of idiomatic expressions. Specifically, by contrasting the two models used for our experiments, we can see a pattern analogous to human understanding of idiomatic expressions. The model which has seen significantly more data, likely including some or all of the expressions in our dataset, is able to disambiguate idiomatic expressions with much better efficacy. Similarly, a person or model which has almost certainly not seen the same idiomatic expressions is largely unable to disambiguate them due to the gap between their lexical foxrm and semantic \textit{I can't think of the word I want here}.

The explicit goal of this paper is to motivate the need for and inspire approaches to designing systems that assist language learners. A summarization of our contributions is as follows: 
\begin{enumerate}[noitemsep,topsep=0pt,leftmargin=*]
    \item We introduce a new cross-lingual idiomatic expression explanation task.
    \item We release, to our knowledge, the first cross-lingual dataset for this task containing English idiomatic expressions as well as explanations of their meanings in English, Spanish, and German.\footnote{Data and code available at: \url{https://github.com/blast-cu/CLIX}. We are not able to publish the Oxford-ME portion due to copyright, but can provide the data upon request.}
    \item We evaluate a set of state-of-the-art models for the task and show that, while automated metrics paint a less-favorable picture, the outcome of few-shot LLMs is judged positively by native speakers of the target language. 
    \item We perform a detailed error analysis and highlight key areas that need improvement before we can confidently use automated cross-lingual explanations for non-standard language in educational applications.
\end{enumerate}

\section{Related Work}

\paragraph{Disambiguation of Non-Standard Language} Non-standard language takes many different forms ranging from slang to metaphor to idiom. To disambiguate some more general forms of non-standard language, \citet{ni_learning_2017} proposed a hybrid word-character sequence-to-sequence model that can disambiguate language found on social media. \citet{kurfali-ostling-2020-disambiguation} show that sufficiently different contextual embeddings can be found for both literal and idiomatic instances of a multi-word expression, thus allowing for disambiguation of the expressions in context. \citet{qiang2023chinese}, on the other hand, try to directly paraphrase Chinese idioms such that they can replace the idiom using an infill strategy similar to the cloze task. These approaches are all tested in monolingual settings. The cross-lingual settings for which a solution might benefit language learners remain less explored.

The cross-lingual version of idiomatic expression disambiguation has traditionally been approached as a translation task. Often, sentences containing idiomatic expressions are mapped to equivalent sentences without such an expression, but models may try to translate them word-by-word \citep{baziotis-etal-2023-automatic}. \citet{fadaee2018examining} explore sentence-level translation to generate an English--German idiom translation dataset, additionally using attention-based model to translate sentences containing idiomatic expressions, requiring in-context disambiguation. \citet{zaninello-birch-2020-multiword} show that data augmentation and external linguistic resources can sizably improve translations of multi-word expressions.
%There has additionally been use of adapters to assist in production of higher quality embeddings for idiomatic expressions with the goal of translation \citep{zeng-bhat-2022-getting}.
%Loss upweighting of sentences possibly containing idiomatic expressions as well as retrieval-augmented models have also been shown to improve translation capabilities, both in scenarios with and without idiomatic expressions \citep{liu-etal-2023-crossing}.
More recently, \citet{Li_Chen_Yuan_Wu_Yang_Tao_Xiao_2024} take a very similar prompting-based approach to us. However, the authors' primary goal appears to be the creation of an ontology, they perform their experiments with languages we do not consider, and they additionally remain focused on a translation-based perspective. We, however, formulate the task more closely to definition generation and explore the capabilities of NLP systems for this purpose in addition to applications in high-stakes scenarios such as education. We believe that this reformulation opens up interesting research avenues, such as investigating whether end-users benefit more from explanations, examples usage, or a combination. We also see new datasets specifically intended for determining whether an idiomatic expression is being used literally or figuratively given some context -- now at the cross-lingual level \citep{sentsova-etal-2025-multicopie}. To the best of our knowledge, however, we are the first to release a cross-lingual dataset specifically for the purpose of explaining these kinds of expressions.

\paragraph{Definition Generation} Since \citet{noraset-definition-2017} introduced this task, new techniques have continued to be introduced to improve performance. \citet{gadetsky_conditional_2018} created embeddings from dictionary definitions, while taking into account the issue of polysemy by allowing their model to recognize that there could potentially be an infinite number of senses for a given word. In contrast, using a novel encoding scheme, \citet{bevilacqua-etal-2020-generationary} are able to autoregressively generate definitions as well as score them in a discriminative manner. \citet{zhang-etal-2023-assisting} generate cross-lingual definitions of words in English, Chinese, and French using prompt learning. They apply a contrastive learning objective to correct pre-trained models' tendency to ignore this task and/or mix languages in the output. In this paper, we focus specifically on idiomatic expressions and their explanations, which allows us to be more flexible and we hypothesize is more challenging for NLP systems.

% we attempt to generate natural language explanations for the idiomatic expressions as we hope to inform a language learner about both the meaning and usage of a given expression.

\section{CLIX}

In this section, we define the CLIX task and describe the process to create a dataset of idiomatic expressions in English with their corresponding explanations in Spanish and German. We present statistics for the resulting dataset and describe the models that we will use to generate predictions.

\subsection{Task Definition} 

Our goal is to provide an explanation of the meaning of an idiomatic expression in their native language of a language learner with the assumption that it would be more understandable than an explanation in the language they are attempting to learn. Following prior work on definition generation, we formalize our task as a text-to-text generation task. Given an idiomatic expression $I=\{i_1,i_2,i_3,...\}$ and some optional context $C=\{c_1,c_2,c_3,...\}$ in the source language $L_S$, our goal is to produce an explanation $E=\{e_1,e_2,e_3,...\}$ in the target language $L_T$. While this formulation is essentially identical to that of definition generation, we opt to call our generated text explanations. While minute, the difference we wish to highlight is that we feel a definition is generally more static as opposed to an explanation which may be tailored to an individual or particular context. Explanations might also include additional contextual information that can be realized in either speech or writing. While we do not require any of these factors to be present in our explanations, because our focus is on language learners and educational contexts, we wish to allow flexibility in how explanations are presented to a user in current and future exploration of this task. Furthermore, EPIE \citep{saxena_epie_2020} and our modified version EPIE-ME do not only contain static definitions. We see samples such as \textit{cheek and jowl} which are instead given an explanation: \textit{Positioned very close together. The cheek and the jowl — the lower part of the jaw — are in close proximity to each other on the face.}

%Although machine translation is a well studied area in NLP and models typically perform quite well, our goal being an explanation means we cannot necessarily leave it all up to the model. To demonstrate this we explore an explain-then-translate pipeline approach, fine-tuning for a direct cross-lingual explanation, and a prompt enhancement strategy.

%While previous definition generation tasks incorporate context into the input (e.g., sentences in which the idiom appears), we do not consider this in our task formulation. This is intentional, as we are looking to emulate a scenario in which a learner is looking to find the meaning of a new word or phrase by looking only that constituent up -- e.g., by typing it into an app on their mobile phone.% and not an example sentence. %The same is true in the case of translation, only the confusing portion is used as a query. Additionally, the transformation we are attempting is not at the sentence level, we only wish to disambiguate the target expression.

\begin{table}
    \centering
    \small
    \begin{tabular}{lccc}
        \hline
        \toprule
        EPIE-ME & Train & Development & Test \\ 
         \midrule
        Idioms & 278 & 150 & 200 \\
        I Avg Len & 3.04 & 3.38 & 3.16 \\
        EN Avg Len & 18.23 & 17.16 & 17.16 \\
        ES Avg Len & 19.30 & 17.98 & 9.63 \\
        DE Avg Len & 18.52 & 17.41 & 9.10 \\
        \toprule
        Oxford-ME & Train & Development & Test \\ 
         \midrule
        Idioms & 4352 & 933 & 933 \\
        I Avg Len & 3.74 & 3.77 & 3.68 \\
        EN Avg Len & 9.04 & 9.08 & 9.08 \\
        ES Avg Len & 9.34 & 9.47 & 9.11 \\
        DE Avg Len & 9.21 & 9.21 & 9.06 \\
        \bottomrule
    \end{tabular}
    \caption{Dataset Statistics. I Refers to Idiom, EN, ES, and DE Refer to English, Spanish, and German Explanations}
    \label{tab:dataset}
\end{table}

\subsection{Dataset} 

%Because parallel datasets are unavailable for cross-lingual idiom definition generation \citep{fadaee2018examining}, 
We construct -- to the best of our knowledge -- the first two datasets for cross-lingual explanation generation of idiomatic expressions, EPIE-ME and Oxford-ME (EPIE and Oxford with Multilingual Explanations), each containing explanations from languages belonging to two separate branches of the Indo-European family, Spanish and German.
%we create our own. 
We build on the EPIE corpus~\cite{saxena_epie_2020}, composed of 25,206 sentences containing instances of 717 formal and static idiomatic expressions with definitions, and example usages.\footnote{Formal idiomatic expressions can undergo lexical changes, while static ones cannot.} We select this dataset because it was in-part originally meant for detection of idiomatic expressions. As a result, it also contains expressions that are situationally idiomatic, a useful quality for judging explanation ability. We filter out expressions for which there are no gold explanations or in-context examples available, leaving us with 628. We additionally use the Oxford Dictionary of Idioms, 4th Edition \citep{Oxford}, comprising of 6,218 English idiomatic expressions and their explanations. This is generally considered the most comprehensive resource for idiomatic expressions in English. The Oxford data only comes with sentence-level context for 1,780 of the expressions (28.6\%), so we prompt Llama 3.1 \citep{dubey2024llama3herdmodels} to generate sentences using missing expressions in the manner found in Tab.~\ref{tab:prompts}. For each unique expression in the corpus, we include an explanation in English, Spanish, and German. We obtain non-English explanations by feeding the gold English explanations to Google Translate. In Tab.~\ref{tab:dataset} we show the train/dev/test splits and the average word length of the idioms and explanations for both datasets. Splits are generated by random sample.

\paragraph{Manual Corrections} For the test sets, the English explanations and the automated translations are closely inspected and manually corrected by native speakers of all three languages to ensure they are entirely correct and not just the output of the automated translation. Any confusion about the meaning of a particular expression was referred to the native English speaker for review. These annotators are all computational linguists fluent in English and working in English speaking countries. 

\paragraph{Idiom Categorization} For EPIE-ME, we further enhance our idioms with thematic categories inspired by findings suggesting that classifying idioms facilitates the learning process~\cite{10.1093/applin/21.4.553}. To do this, we build on the taxonomy developed by \citet{RafatbakhshElaheh2019Atcs}, which suggests a list of 81 themes to classify English idioms (See full list with frequencies in App. \ref{app:themes}). For any idioms already assigned themes by \citet{RafatbakhshElaheh2019Atcs} (a little over 20\%), we use the assigned theme. To tag the remainder of our dataset using this taxonomy, we prompt GPT-3.5 Turbo \citep{gpt-3.5} to provide us with the three most appropriate categories over three runs where the categories are provided in random order. We then eliminate duplicates in the resulting nine predictions and correct them by hand. The manual corrections were done by two English speaking annotators working together to resolve conflicts.

\paragraph{Estimating Noise in Non-Corrected Examples} Finally, we examine the amount of noise in the explanations without manual correction. We do so by calculating the edit distance \citep{Levenshtein1965BinaryCC} between the original output of Google Translate and the corrected version produced by our annotators over our verified test set. %\footnote{For Oxford, we only verify the Spanish translations.}, 
We further normalize by length following \citet{tashima-etal-2018-fault}. We see that the amount of correction required is generally low, the largest amount being required on average for the German EPIE-ME explanations with a normalized edit distance of 0.283, while Spanish had an edit distance of 0.159. More information can be found in Appendix \ref{app:noise}.

\section{Learning and Inference Strategies}\label{strategies}

In this section, we outline the different learning and inference strategies that we experiment with to generate cross-lingual explanations of idiomatic expressions. We experiment with fine-tuned and few-shot generation models. Strategies include: whether performing the task end-to-end or decomposing it into explanation and translation, the type of contextual information used to enhance the input prompts, and the number of demonstrations used for LLM-based solutions. All prompt templates can be found in App.~\ref{app:prompts}

\subsection{Explanation Strategies}

\paragraph{Direct Explanation} To obtain an explanation for the benefit of a potential language learner, we fine-tune and/or prompt models for an explanation in what we assume to be the L1 of the learner. All inputs, however, are in English. Our training pairs in the fine-tuning setup consist of an input $X_i=[T_{i,{L_T}};I_i]$ and label $Y_i=[E_{i,{L_T}}]$, where $T_{i,{L_T}}$ is set of tokens informing the model that its task is to produce an explanation in a target language $L_T$, $I_i$ is an idiomatic expression, and $E_{i,{L_T}}$ is a gold explanation of the idiomatic expression. This model can then be formulated as:

\begin{equation}\label{eq:model}
    P(Y_k|X_k)=\prod_t(y_k|y_{0:k-1},X_k;\theta)
\end{equation}

% A typical input in its most basic textual form is \textit{Explain in $L_T$ the idiomatic expression $I$} for both the fine-tuning and prompting setups.

\paragraph{Pipeline} In this configuration, we aim to generate our final explanation in two stages. The first stage is to generate an explanation in English. In our fine-tuning setup, we use T5 \citep{2020t5} for this step. Where mT5 \citep{xue-etal-2021-mt5} was explicitly designed to learn many languages, T5 was only trained to translate between a small set, and most of its training data was English. Thus, we feel T5 is more likely to perform well on this task. The model definition remains the same as in Eq.~\ref{eq:model}. However, the input pair is now $X_i=[T_i;I_i]$ and $Y_i=[E_{i,L_S}]$. In this case, our task sequence $T_i$ no longer provides any language information, and our explanation $E_{i,L_S}$ remains in the source language (in our case, English).
% A lexical input to this first step is typically of the form: \textit{Explain the idiomatic expression $I$}.

The second stage of our pipeline requires a translation of the output of the previous step. Because T5 was trained for the task of translation to/from German but not to/from Spanish, any comparison between results would be unfair. While we considered the use of mT5 in this second stage, its performance on translation tasks is already well understood \cite{xue-etal-2021-mt5}. As a result, and to additionally be consistent with our choice in the curation of our dataset, we use Google Translate for the second step of our pipeline in the fine-tuning scenarios. In the experiments performed with GPT and Llama, however, the models are asked to translate their own output.

To validate our assumptions we provide results for a full-mT5 pipeline where we fine-tune versions for each of the explanation and translation tasks.

\begin{table}
    \small
    \centering
    \resizebox{\columnwidth}{!}{%
    \begin{tabular}{lcc}
    \toprule
    Theme & Spanish & German \\
    \midrule
        Language, speech, and conversation & 71.66 & 72.07 \\
        Happiness, pleasure, and enjoyment & 76.88 & 79.41 \\
        Health and illness & 70.20 & 71.46 \\
        Argument and conflict & 78.97 & 77.62 \\
        Cooperation & 77.61 & 79.01 \\
        Experience & 71.99 & 68.65 \\
        Duty and responsibility & 68.37 & 69.14 \\
        Success & 77.92 & 78.95 \\
        Haste and speed & 72.77 & 76.57 \\
        Anger and annoyance & 87.98 & 73.12 \\
    \bottomrule
    \end{tabular}
    }
    \caption{Performance of Best Model as Measured by Sentence Similarity on EPIE-ME Data per Theme for Top 10 Themes}
    \label{tab:perf_themes}
\end{table}

\subsection{Contextual Enhancements}\label{enhancements}
We explore different ways to add contextual information into the input prompt. This is done only for LLM models: Llama and GPT.

\paragraph{Sentence-Level Context} Because idiomatic expressions encode information that is potentially unrelated to their linguistic form, many researchers have used sentence-level context to assist with identification, disambiguation, and translation of idiomatic expressions \citep{fadaee2018examining, kurfali2020disambiguation, saxena_epie_2020, liu-etal-2023-crossing, Li_Chen_Yuan_Wu_Yang_Tao_Xiao_2024}. We experiment with including example sentences (one usage example per instance) containing the idiomatic expression to be explained at the end of our prompts in an effort to provide more information to the model.

\paragraph{Categorical Enhancements} In this setting we explore how the addition of small hints can help these models. We add to our prompt a category that an idiomatic expression might belong to. 
% resulting in a prompt of the form \textit{Explain in $L_T$ the idiomatic expression $I$ in the context of $C$} where $C$ is the category the expression belongs to.
%In our zero-shot experiments, we aim to determine whether the addition of a known category is of any benefit to the model. In our few-shot experiments, we do not inform the model of the category, but rather investigate whether selecting in-context learning examples from random, induced, or known categories is most effective. 
All categories we use come from the \citet{RafatbakhshElaheh2019Atcs} taxonomy.

\begin{table*}[!t]
    \centering
    \small
    \begin{tabular}{l|lll|lll}
    \toprule
        \multirow{2}{*}{\textbf{Model}} & \multicolumn{3}{c|}{\textbf{EPIE-ME}} & \multicolumn{3}{c}{\textbf{Oxford-ME}} \\
        & Overall & Spanish & German & Overall & Spanish & German \\
        \midrule
        mT5 Direct & 38.06 & 36.32 & 39.80 & 43.21 & 43.56 & 42.86 \\
        mT5 Pipeline & 37.13 & 37.13 & 37.81 & 42.58 & 42.40 & 42.77 \\
        T5 Pipeline & 43.54 & 43.19 & 43.89 & 46.09 & 45.96 & 46.22 \\
        \midrule
        Llama Direct Zero-Shot & 59.39 & 60.13 & 58.66 & 55.32 & 56.01 & 54.62 \\
        Llama Pipeline Zero-Shot & 60.16 & 60.48 & 59.83 & 55.01 & 55.10 & 54.91 \\
        \midrule
        GPT Direct Zero-Shot & 65.06 & 65.96 & 64.16 & 61.03 & 62.02 & 60.03 \\
        GPT Pipeline Zero-Shot & 69.60 & 70.12 & 69.07 & 66.10 & 66.28 & 65.91 \\
        \midrule
        GPT Direct 5-Shot & 71.15 & 71.42 & 70.87 & 66.13 & 66.9 & 65.36 \\
        GPT Pipeline 5-Shot & \textbf{71.84} & \textbf{72.04} & \textbf{71.63} & \textbf{68.54} & \textbf{68.87} & \textbf{68.20} \\
        \bottomrule
    \end{tabular}
    \caption{Best Model Results by Sentence Similarity for Direct and Pipeline Strategies. LLM Results are Averaged Across 3 Runs on the Test Set. Scores for English Explanations in the First Half of the Pipeline can be Found in Tab.~\ref{tab:pre_translation}}
    \label{tab:main_results}
\end{table*}

\subsection{Few-Shot Prompting}\label{few-shot}

We explore the effects of few-shot prompting with both Llama and GPT. We implement scenarios where the models are provided with 1, 3, or 5 examples from which they can learn in-context. While there is some expectation that more demonstrations will increase performance, we also aim to determine whether or not the same categorical information used for context enhancement could be used to strategically select examples and further increase performance. We compare random selection to a scenario in which we induce the category label from the LLM and one in which we assume that the category label is known.

To induce a label, we ask the model to choose the most suitable label from our list of categories. If the induced label does not exactly match an existing label, we add it to our list and treat it as an existing label for only the instance in which we observe this idiomatic expression, and from this point forward, the procedure is the same as if we assume the category is known. Once we have a known category label (either given or acquired by induction), meaning we are given some idiomatic expression $I_i$ with category $C_i$, we form an initial selection of $2*k$ idiomatic expressions with category label $C_j$ where $C_i=C_j$ and $k$ is the number of shots. If there are too few expressions to form this selection, we draw from the most similar categories until we obtain $2*k$ examples. Similarity between categories is calculated as the similarity between the labels by the same SBERT \citep{reimers-2020-multilingual-sentence-bert} model used for our evaluation in Sec.~\ref{experiments}. Once these $2*k$ examples are obtained, we take a final sample of $k$ from this set. This sampling strategy is necessitated by the small size of our dataset, where we may not have even a single example in our training set with the same category label as one in our validation or test set. The first step of selecting $2*k$ initial examples helps reduce variance between instances where there are enough examples where the categories match and instances where there are too few.

\begin{table}[!t]
\centering
\small
\begin{tabular}{l|cc}
\toprule
\textbf{Enhancement} & {\textbf{GPT}} & {\textbf{Llama}}\\
\midrule
Direct & \textbf{63.91} & \textbf{60.45} \\ 
+ SL & 61.36 & 59.73 \\
+ Cat & 61.96 & 59.44 \\
+ SL + Cat & 61.61 & 58.98 \\
\midrule
Pipeline & \textbf{66.98} & 61.32 \\ 
+ SL & 66.22 & \textbf{64.00} \\
+ Cat & 65.64 & 60.53 \\
+ SL + Cat & 66.36 & 63.04 \\
\bottomrule
\end{tabular}
\caption{Zero-Shot Results on EPIE-ME by Contextual Enhancement. SL: Sentence-Level, Cat: Categorical Information.}\label{tab:dev_llm}
\end{table}

\section{Experiments}\label{experiments}

We compare results of several models across a range of strategies outlined in Sec.~\ref{strategies}. We aim to investigate what makes certain models better at understanding idiomatic expressions and what types of supplemental information are beneficial for in-context learning within this domain. We discuss the addition of contextual information outlined in Sec.~\ref{enhancements}. We also more closely examine methods of selecting good demonstrations in few-shot prompting scenarios when levels of information for the newly-observed idiomatic expression vary, as described in Sec. \ref{few-shot}.

\paragraph{Models and Metrics} In our experiments, we benchmark the performance of two sequence-to-sequence models, T5 \citep{2020t5} and mT5 \citep{xue-etal-2021-mt5}, as well as two LLMs, Llama3.1\footnote{Llama3.1 8B Instruct} \citep{dubey2024llama3herdmodels} and GPT-3.5 Turbo\footnote{Model Version: \texttt{gpt-3.5-turbo-0125}} \citep{gpt-3.5}. To evaluate our approaches we primarily follow \citet{zhang-etal-2023-assisting}. To determine the quality of the generated explanations, we calculate their similarity to the gold explanations in the English using a multilingual version of SBERT \citep{reimers-2020-multilingual-sentence-bert}. We use this metric as it is more likely to assess the meaning of the expressions. Finally, in Tab.~\ref{tab:other_metrics} we include BLEU and ROUGE scores calculated with our gold translated explanations for the best models using SacreBLEU \citep{post-2018-call} and ROUGE-L.

\subsection{Results and Discussion}

Our main results are described in Tab.~\ref{tab:main_results}. In the case of mT5 and T5, the quality of the results is generally unimpressive, demonstrated by the semantic similarity scores hovering around 40. However, we find that a T5-based pipeline strategy increased performance by more than 10\%. Upon manually inspecting the outputs, in both cases the output is solely in the indicated language, but we did not see a significant difference in the fluency between the different strategies. There were, however, cases where the model would continuously repeat a key word or phrase in its response. Examples include \textit{Una persona típicamente \textbf{atractiva}, \textbf{atractiva} o \textbf{atractiva}} and \textit{Un estado de \textbf{desorientación} o \textbf{desorientación}}. In the case of the models trained with Oxford-ME data, we only see a 10\% increase in performance, indicating that models trained on the smaller EPIE-ME dataset remain competitive with those trained on 10 times the amount of data.

%more data does indeed help the models, but not enough so to perform well above those trained on less data.

In our experiments using LLMs, we see significant improvements in performance, reaching a similarity of nearly 70 with the best zero-shot strategy. This result is unsurprising, given that GPT and Llama have been exposed to significantly larger amounts of data, likely including many in context occurrences of the idioms in our dataset. Consistent with the results from the sequence to sequence models, pipeline strategies seem to be most effective. While the final difference in score between the 5-shot direct and pipeline approaches is very small, we see a larger gap in the zero-shot setting. We believe that the increased gains in the direct setup are due to the direct scenario requiring more internal reasoning steps, where the demonstrations appear to help make up the difference. In contrast, the pipeline method appears to approach an upper bound in similarity scores sooner. In the case of the similar scores when using Llama in a zero-shot setting, we cannot find an exact reason. However, the fact that the pipeline approach universally performs better indicates that it is most effective. For the Oxford-ME dataset, we observe that the improvement of LLMs over fine-tuned models is smaller. We attribute this to the gold explanations in the Oxford-ME dataset being about 33\% shorter, penalizing the often verbose responses of the LLMs.

In the final results of our experiments, we found that while additional context benefited Llama in certain cases, it did not help GPT. While we are unable to pinpoint why it was unhelpful, we hypothesize that it is a result of the model both being more capable and the lack of a direct link between the contextual enhancement and the gold explanation. This absence is particularly noticeable in relation to experiments where the model is provided categorical information. Dev set performance can be found in Tab.~\ref{tab:dev_llm}. Lastly, we see consistently higher performance by GPT when compared to Llama across all metrics. This is likely due to our use of the -- small for an LLM -- 8 billion parameter version of Llama, and we do not necessarily expect to see the same trend when using a larger version.

When comparing our zero and few-shot experiments, we can clearly see an increase in performance as the number of shots increases. However, we find that differing selection strategies for our examples in the few-shot scenarios do not help whatsoever. The model that performed the best used the random selection strategy. This indicates to us that either the model already represents the idiomatic expressions very well, or the relatedness of examples chosen for in-context learning are largely unimportant for this task. Results calculated on our dev set can be found in Tab.~\ref{tab:dev_few_shot}.

Finally, we perform an evaluation of the performance per theme (See Tab.~\ref{tab:perf_themes}) using the best performing model on our test set. We can see that for the majority of our most common themes, this model is performing above expectations with respect to the overall results in Tab.~\ref{tab:main_results}. Intuitively, the model is able to better explain idiomatic expressions from more common domains.

%While prompting LLMs gives us better results, we can see that there is still significant need for improvement before these explanations are usable in target applications. 

% This no longer applies :(
% However, we see that in the pipeline setting the results are much more similar. We attribute this to the nature of LLMs being much larger and trained much more extensively, thus being able to cope with the higher level of complexity in the cross-lingual transformation. We can see that mT5 is able to do the reasoning portion of the task (monolingual explanation) relatively well. Then all the model needs to do is what it is best at, translation.

\begin{table*}[!ht]
\small
     \resizebox{\textwidth}{!}{
     \begin{tabular}{>{\arraybackslash}m{2.5cm}|>{\arraybackslash}m{12.5cm}}
     %\begin{tabular}[ll]
     \toprule
       \textbf{Error Type}  &  \textbf{Examples} \\
       \midrule
       Word repetition and redundant language (Fluency) &  
       \begin{itemize}[leftmargin=*,nosep]
           \item producir \textbf{resultados} positivos o \textbf{resultados}. \textcolor{blue}{[ENG: produce positive \textbf{results} or \textbf{results}]}
           \item \textbf{abandonar} a alguien o algo repentinamente y sin previo aviso, a menudo \textbf{abandonándolos}. \textcolor{blue}{[ENG: to \textbf{abandon} someone or something suddenly and without notice, usually by \textbf{abandoning} them]}
           \item Algo o alguien que es un placer y una \textbf{vista} agradable de \textbf{ver}. \textcolor{blue}{[ENG: something or someone who is a pleasure and a (\textit{literal translation of idiom}) \textbf{sight} worth \textbf{seeing}]}
       \end{itemize}
       \\
       \midrule
       Odd grammar (Fluency) & \begin{itemize}[leftmargin=*,nosep]
           \item Aquellos que lleguen o apliquen primero serán los primeros en \textbf{recibir} o ser atendidos. \textcolor{blue}{[ENG: Those who arrive or apply first will be the firsts to \textbf{receive} or be seen]} -- \textit{uses verb to receive with no direct object.}
           \item Refiriéndose a \textbf{algo} que es \textbf{el} más avanzado o sofisticado de su tipo en un momento dado. \textcolor{blue}{[ENG: Referring to \textbf{something} which is \textbf{the} most advanced or sophisticated of its type in a given time.]} -- \textit{wrong co-referent article for `algo', uses masculine form `el' insead of neutral `lo'. In English they both translate to `the'.}
       \end{itemize}
       \\
       \midrule
       Meaning mismatch (Accuracy) & 
       \begin{itemize}[leftmargin=*,nosep]
       \item \textbf{One's flesh and blood}: Se refiere al cuerpo físico, pero a menudo se usa para enfatizar el aspecto humano de alguien o algo, \textcolor{blue}{[ENG: About the physical body, but often used to emphasize the human aspect of something or someone.]} -- \textit{identifies that the body is used as a metaphor, but ties it to humanity rather than family.}
        %\item \textbf{Play cat and mouse:} Participar en un juego de persecución o engaño, donde una persona o grupo (el gato) intenta atrapar o engañar a otra persona o grupo (el ratón). -- \textit{talks about deceit and persecution, rather than evasiveness.}
       \end{itemize}
       \\
       \midrule
       Literal meaning (Accuracy) & 
       \begin{itemize}[leftmargin=*,nosep]
           \item \textbf{Dice with death:} Participar en una actividad arriesgada o peligrosa que podría resultar potencialmente en la muerte. \textcolor{blue}{[ENG: Participate in an activity that could cause death.]} -- \textit{Makes explicit mention to death, rather than seeing it as language for danger.}
       \end{itemize}\\
       
       \bottomrule

    \end{tabular}
    }
    \caption{EPIE-ME Common Error Types}\label{tab:error-types}
\end{table*}

\subsection{A Note on Translation}

Although there has been success translating idiomatic expressions in the past \citep{fadaee2018examining,zaninello-birch-2020-multiword}, translation alone is not sufficient in educational scenarios. In many past approaches, the task assumes that a model will be provided with at least sentence-level context, which helps models perform well on translation-based tasks. However, there is no guarantee that a live user will give a model an entire sentence or more when seeking to understand an idiomatic expression. Our task tests the robustness of these models to a lack of context. %when a user query may only contain an idiomatic expression they don't understand
In Appendix~\ref{app:translation}, we outline an experiment where data from human evaluators indicates that 42\% of Spanish and 48.5\% of German translations remain unnatural, while good translations will be word-for-word in roughly 50\% of cases and thus cannot be used as a viable method for this task unless for an idiomatic expression in $L_S$ there is an equivalent expression in $L_T$.

\subsection{Data Contamination}

There is always potential for data contamination in LLM experiments. While we cannot make claims about what data the LLMs may have seen in training, we attempt to correlate the explanation generation performance with the number of times an idiom appears online and the likelihood of generating a particular expression from in context examples using an LLM. We find there is \textbf{no correlation} (Spearman's $\rho$ between -0.03 and -0.3) between either proxy measure and model performance. We include a comprehensive report in Appendix~\ref{app:contamination}.

\section{Human Evaluation}\label{human-eval}

Our main goal is to motivate the need for and inspire design approaches for tools that can assist language learners. As a result, we perform manual evaluation to determine how helpful the best model's explanations might be to real people. Following \citet{zhang-etal-2023-assisting}, we measure two qualities of this model's generated explanations, Fluency and Accuracy. For the highest performing model's Spanish generations, we ask native speakers to rate both aspects on a Likert scale ranging from 1-5. The accuracy measure captures the congruence between the generated explanation and the meaning of the idiom. Fluency, however, captures the level of familiarity with the language indicated by the model's response. There can often be a challenge in describing fluency of a text, especially when short. Thus, we create annotation guidelines inspired by the American Council on the Teaching of Foreign Languages proficiency guidelines \citep{actfl2024}, but modified to suit our specific task. Detailed rubrics can be found in App.~\ref{guidelines}. The generations were rated by two annotators independently. Both annotators are computational liguists and native Spanish speakers from different Spanish-speaking regions: Europe and South America. Both annotators are also fluent English speakers. 

Our annotators found that the model performed extremely well in terms of both Fluency and Accuracy, with average scores of 4.70 and 4.78 respectively. To measure inter-annotator agreement, we calculate Krippendorff's $\alpha$. These scores range from -1 to 1, where 1 is perfect agreement, 0 is random agreement and -1 is inverse agreement. We found $\alpha=0.642$ for fluency and $\alpha=0.417$ for accuracy. We found two main reasons for disagreement. First, in some cases, an explanation may begin with a literal interpretation and end with a figurative interpretation. In this case, one of our annotators appeared to penalize for the inclusion of the literal interpretation where the other did not. Second, our raters' native variants of Spanish are different, resulting in differences in understood meaning or connotation of certain words and phrases. However, we do not consider difference in variant a significant issue as it pertains to our overall results. Since the generative models examined in this paper are trained on the same generalized Spanish data that the embedding model is trained on (all ignoring the existence of any variants), the scores given by our annotators ought to be lower than an automated evaluation rather than higher (assuming the automated evaluation is infallible).

\subsection{Error Analysis}

During human evaluation, our raters pointed out common errors affecting both fluency and accuracy. Issues with fluency come from the model's tendency to repeat words that may be key to the meaning of the idiomatic expression or make grammatical mistakes. Problems with accuracy arise in cases where the model is either too literal or misidentifies the specific type of entity or domain that an expression refers to. Our annotators also observed that the model might introduce information not necessarily related to the meaning of the expression. Rather, it might indicate the origin of the idiomatic expression, or an indicated stance of the speaker. We see this as a positive addition to the explanation in educational contexts, but there is inconsistency and we neither ask for nor measure this phenomenon in our experiments. In the context of our educational goals, these findings present a few key issues that we would not wish to see arise in classroom settings. Despite our high fluency and accuracy scores reported in Sec.~\ref{human-eval}, there is clearly still room to grow before these models are given real responsibility in educational domains.

\section{Conclusion and Future Work}

This work represents a first step towards generating cross-lingual explanations of idiomatic expressions. We introduced two datasets, one consisting of 628 English idioms and another containing 6218 in English idioms, each with their explanations in English, Spanish and German, and showed that state of the art systems perform well when generating cross-lingual explanations, but not well enough in certain aspects for our ideal target domain of education. For our learned models, we wish to devise more effective strategies that incorporate some notion of directly unwinding figuration and explicit focus on a complete explanation that delivers information so that a learner attains some requisite understanding.

We also showed a disconnect in automated measurements of performance and human judgements with relation to our task. Because the model only generated small samples of text, there is no perfect way to evaluate its overall fluency within a specific language. Although sentence similarity is one way to evaluate accuracy, it does not necessarily focus on whether the core of the idiomatic expression is being represented in the explanation, only how the model represents certain pieces of text.

Finally, although outside the scope of this paper, we can see that some idiomatic expressions are more unconventional than others, but we have no obvious way of measuring or directly coping with this phenomenon. Given the examples \textit{building bridges}, \textit{break a leg}, and \textit{kick the bucket}, clearly the first is more clearly linked to its meaning by metaphor than the latter two, and might even be more easily disambiguated with less context because of the kind of words required to surround it. This is something we would like to measure in the future. Specifically, assuming idiomatic expressions live on this gradient of metaphor, are expressions on one end of the spectrum easier for models to disambiguate than those on the other?
%we would like to improve our evaluation to consider alternative metrics, such as human judgements of clarity and usefulness. Additionally, we would like to perform a more extensive analysis on the difficulty of idioms by their type, considering other linguistically informed taxonomies. This would allow us to investigate if there are any parallels between expressions that are harder for humans to understands and those that models struggle to disambiguate.

\section{Limitations}

Given that we cannot make Oxford-ME public, the primary limitation of our paper is the small size of our publishable dataset. However, we are able to provide this data to other researchers upon request. %The fact that we only have access to 628 idioms (resulting in 1,256 examples total, one for each target language) limits our ability to learn the \textcolor{red}{best possible generation model}.
%To combat this issue, we wish to augment our dataset with additional idioms, using resources such as the Oxford Dictionary of Idioms, which includes a total of 1,506 idioms.
Additionally, we are limited in the number of languages we support. We have only one source language of idiomatic expressions, English. There is a reasonable chance that the models we have experimented with might perform worse in experiments conducted with languages that have smaller amounts of available data. Furthermore, we only attempt cross-lingual explanation into two languages, Spanish and German, which are similarly well supported with data. Finally, the metrics we use in this paper are simply not able to fully capture the phenomenon we wish to measure. We hope to motivate and/or participate in the development of more informative metrics that best suit a task such as ours.

\section*{Acknowledgments}

We thank members of the BLAST and NALA labs for their feedback. This work utilized the Blanca condo computing resource at the University of Colorado Boulder. Blanca is jointly funded by computing users and the University of Colorado Boulder.

%The second limitation of our paper is that we limit our evaluation to automated metrics. In future work, we would like to incorporate a comprehensive human evaluation to understand the main points of failure of these system, and inform future strategies.

%\section*{Acknowledgements}
% Commented out until camera-ready

\bibliography{anthology,custom}

\appendix

\section{Hyperparameters and Other Tables}
\label{app:dev_set}

For T5 and mT5, we use a learning rate of 1e-5, the AdamW optimizer, a batch size of 32, and run for 1000 epochs with early stopping. For GPT and Llama, we also explore two different configurations to constrain the length of our explanation to a maximum of one or two sentences. %Our best model, however, was not subject to a length constraint. 
Additionally, for GPT 3.5 Turbo we set the temperature to 0.5 for all scenarios, and Llama was left with its default temperature. All hyperparameters were chosen using the development set.
% (results in App.~\ref{app:dev_set})

% We include results of our fine-tuned models for the development set in Tab.~\ref{tab:dev_fine-tune}.

%\textbf{TODO}
\begin{table}[!ht]
    \small
    \centering
    \resizebox{\columnwidth}{!}{
        \begin{tabular}{l|ll|ll|ll}
            \toprule
            \multirow{2}{*}{\textbf{Model}} & \multicolumn{2}{c}{\textbf{Overall}} & \multicolumn{2}{c}{\textbf{Spanish}} & \multicolumn{2}{c}{\textbf{German}} \\
            & BLEU & ROUGE & BLEU & ROUGE & BLEU & ROUGE \\
            \midrule
            mT5 Direct & 0.37 & 9.94 & 0.34 & 10.59 & 0.47 & 9.34 \\
            mT5 Pipeline & 0.09 & 4.90 & 0.18 & 9.71 & 0.29 & 7.80 \\
            T5 Pipeline & 0.89 & 11.95 & 1.03 & 13.09 & 0.74 & 10.73 \\
            \midrule
            Llama Direct Zero-Shot & 1.00 & 12.31 & 1.12 & 14.19 & 0.87 & 10.48 \\
            Llama Pipeline Zero-Shot & 1.40 & 14.69 & 1.67 & 16.54 & 1.08 & 12.88 \\
            \midrule
            GPT Direct Zero-Shot & 3.58 & 22.1 & 3.66 & 24.39 & 3.50 & 19.87 \\
            GPT Pipeline Zero-Shot & \textbf{4.59} & \textbf{26.41} & \textbf{5.56} & \textbf{28.21} & 3.57 & \textbf{24.63} \\
            \midrule
            GPT Direct 5-Shot & 4.04 & 23.23 & 4.22 & 25.09 & \textbf{3.86} & 21.34 \\
            GPT Pipeline 5-Shot & 4.19 & 24.76 & 4.88 & 26.38 & 3.47 & 23.13 \\
            \bottomrule
        \end{tabular}
    }
    \caption{Best Model Results on EPIE-ME Using BLEU and ROUGE for Direct and Pipeline Strategies. LLM Results are Averaged Across 3 Runs on the Test Set.}
    \label{tab:other_metrics}
\end{table}

\begin{table}[!ht]
\centering
\small
\begin{tabular}{lcc}
\toprule
Language & T5 & mT5\\
\midrule
Overall & 48.4 & 41.60 \\ 
Spanish & 48.23 & 41.49 \\ 
German & 48.57 & 42.02 \\ 
\toprule
\multicolumn{3}{c}{Pre-Translation} \\
\midrule
English & 44.83 & 41.14 \\
\bottomrule
\end{tabular}
\caption{EPIE-ME T5 \& mT5 Dev Set Results}\label{tab:dev_fine-tune}
\end{table}

\begin{table}[!ht]
\centering
\small
\begin{tabular}{l|ccc}
\toprule
\textbf{Model} & \textbf{Sim} & \textbf{BLEU} & \textbf{ROUGE} \\
\midrule
mT5 & 37.45 & 0.31 & 11.82 \\
T5 & 42.04 & 0.46 & 13.58 \\
\midrule
Llama Zero-Shot & 57.66 & 1.82 & 18.11 \\
GPT Zero-Shot & 71.05 & 5.59 & 29.87 \\
\midrule
GPT 5-Shot & 72.57 & 4.7 & 28.1 \\
\bottomrule
\end{tabular}
\caption{EPIE-ME Pipeline Model Performance Before Translation (EN output, Test Set)}\label{tab:pre_translation}
\end{table}

\begin{table}[!ht]
\centering
\small
\resizebox{\columnwidth}{!}{%
\begin{tabular}{l|ccc|ccc}
\toprule
\multicolumn{7}{c}{\textbf{GPT}}\\
\midrule
 & \multicolumn{3}{c|}{\textbf{Direct}} & \multicolumn{3}{c}{\textbf{Pipeline}} \\
\midrule
Shots & 1 & 3 & 5 & 1 & 3 & 5 \\
\midrule
Known & 65.86 & 67.50 & \textbf{67.75} & 68.26 & 68.31 & 68.62 \\ 
Induced & 65.13 & 66.16 & 67.52 & 67.77 & 68.09 & 68.57 \\
Random & 65.25 & 67.08 & 67.45 & 67.81 & 68.59 & \textbf{68.79} \\
\bottomrule
\end{tabular}}
\caption{EPIE-ME Few-Shot Results by Demo Selection Strategy}\label{tab:dev_few_shot}
\end{table}

\pagebreak
\section{Human Evaluation Guide}\label{guidelines}
We provide the rubrics given to our annotators for rating both fluency and accuracy in Tabs.~\ref{tab:fluency_rubric} and~\ref{tab:acc_rubric}. The fluency rubric is inspired in-part by \citet{actfl2024}.

\begin{table}[!ht]
\centering
     \resizebox{\columnwidth}{!}{\begin{tabular}{|>{\arraybackslash}m{1cm}|>{\arraybackslash}m{10cm}|}
     %\begin{tabular}[ll]
     \hline
       \textbf{Rating}  &  \textbf{Description} \\
       \hline
       1  &  \textbf{Little to No Understanding:} 
       \begin{itemize}[leftmargin=*,nosep]
           \item Has a severely limited vocabulary, acquiring only a handful of words in the language.
           \item No understanding of grammatical constructs.
           \item Is not in the correct language.
       \end{itemize}
       \\
       \hline
       2 & \textbf{Beginner Language Learner:} 
       \begin{itemize}[leftmargin=*,nosep]
           \item Still includes some words from their native language.
           \item Basic vocabulary - only uses words a child would understand.
           \item Sentence structure that mimics the native language.
       \end{itemize}
       \\
       \hline
       3 & \textbf{Intermediate Language Learner:} 
       \begin{itemize}[leftmargin=*,nosep]
           \item Understanding of standard grammatical structures.
           \item Inefficient communication of ideas - vocabulary is clearly more sizable, but word choice is somewhat odd or ineffective.
           \item Perhaps gets to their point in a roundabout fashion.
         \end{itemize}
           \\      
       \hline
       4  & \textbf{Advanced Speaker:} 
       \begin{itemize}[leftmargin=*,nosep]
           \item Strong vocabulary.
           \item Strong understanding of tense morphology and usage.
           \item Grammar is near perfect - capable of more complex constructs.
         \end{itemize}  \\
       \hline
       5 & \textbf{Fluent/Native Speaker:} 
       \begin{itemize}[leftmargin=*,nosep]
           \item Language production as expected of someone with a complete mastery of the language.
           \item Communicates efficiently and precisely (inclusion of examples to provide clarity should not be seen as inefficient communication).
           \item Produces language as if communicating with another fluent speaker.
         \end{itemize}  \\
       \hline
       
    \end{tabular}}
    \caption{Fluency Rubric}\label{tab:fluency_rubric}
\end{table}

\begin{table}[!ht]
    \centering
     \resizebox{\columnwidth}{!}{\begin{tabular}{|>{\arraybackslash}m{1cm}|>{\arraybackslash}m{10cm}|}
     %\begin{tabular}[ll]
     \hline
       \textbf{Rating}  &  \textbf{Description} \\
       \hline
       1  &  \textbf{Completely Wrong:} 
       \begin{itemize}[leftmargin=*,nosep]
           \item Language may also be too incoherent to determine.
       \end{itemize}
       \\
       \hline
       2 & \textbf{Literal Interpretation:} 
       \begin{itemize}[leftmargin=*,nosep]
           \item Does not understand the figurative nature of the expression, and tries to reference the lexical form when inappropriate (i.e., mentioning chickens in reference to "putting all of your eggs in one basket").
       \end{itemize}
       \\
       \hline
       3 & \textbf{Domain Matches but Not Meaning:} 
       \begin{itemize}[leftmargin=*,nosep]
           \item Identifies the general context in which the expression is to be used, but the meaning is incorrect (i.e., has some sense of related topics, but cannot identify sentiment or value).
         \end{itemize}
           \\      
       \hline
       4  & \textbf{Good Sense of Meaning:} 
       \begin{itemize}[leftmargin=*,nosep]
           \item Near-complete understanding of the expression or the value which it communicates, still referencing the main ideas found in the gold explanation.
         \end{itemize}  \\
       \hline
       5 & \textbf{Perfectly Captures Meaning:} 
       \begin{itemize}[leftmargin=*,nosep]
           \item Very high congruence with the gold explanation.
         \end{itemize}  \\
       \hline
       
    \end{tabular}}
    \caption{Accuracy Rubric}\label{tab:acc_rubric}
\end{table}

\section{Themes}\label{app:themes} We include the full list of themes from \citet{RafatbakhshElaheh2019Atcs} and their frequencies within our dataset (in descending order) in Tabs. \ref{tab:all_themes_1} and \ref{tab:all_themes_2}.

\begin{table}[!ht]
    \footnotesize
    \centering
    % \resizebox{\columnwidth}{!}{%
    \begin{tabular}{lc}
    \toprule
    Theme & Num. Idioms \\
    \midrule
        Language, speech, and conversation & 29 \\
        Happiness, pleasure, and enjoyment & 22 \\
        Health and illness & 22 \\
        Argument and conflict & 20 \\
        Cooperation & 19 \\
        Experience & 19 \\
        Duty and responsibility & 18 \\
        Success & 16 \\
        Haste and speed & 15 \\
        Anger and annoyance & 14 \\
        Hope and optimism & 14 \\
        Caution & 13 \\
        Expense & 13 \\
        Futility & 13 \\
        Honesty & 13 \\
        Power & 13 \\
        Time & 13 \\
        Anxiety and worry & 12 \\
        Crime and punishment & 12 \\
        Work and employment & 12 \\
        Appearance & 11 \\
        Misfortune and adversity & 11 \\
        Thoroughness & 11 \\
        Change & 10 \\
        Chaos and disorder & 10 \\
        Love & 10 \\
        Preparation and readiness & 10 \\
    \bottomrule
    \end{tabular}
    \caption{EPIE-ME Idiomatic Expression Count ($\geq 10$) per Theme}
    \label{tab:all_themes_1}
\end{table}

\begin{table}[!ht]
    \footnotesize
    \centering
    % \resizebox{\columnwidth}{!}{%
    \begin{tabular}{lc}
    \toprule
    Theme & Num. Idioms \\
    \midrule
        Action & 9 \\
        Deception and lying & 9 \\
        Excess and extravagance & 9 \\
        Poverty & 9 \\
        Reputation and fame & 9 \\
        Travel and transport & 9 \\
        Danger & 8 \\
        Age & 7 \\
        Certainty & 7 \\
        Doubt and uncertainty & 7 \\
        Intelligence and knowledge & 7 \\
        Secrecy & 7 \\
        Beauty & 6 \\
        Critics and criticism & 6 \\
        Foresight and the future & 6 \\
        Forgiveness and reconciliation & 6 \\
        Opportunity & 6 \\
        Self-interest & 6 \\
        Courage & 5 \\
        Crisis & 5 \\
        Family & 5 \\
        Fools and foolishness & 5 \\
        Surprise & 5 \\
        Unhappiness and disappointment & 5 \\
        Bribery, corruption, and extortion & 4 \\
        Indecision and prevarication & 4 \\
        Laziness & 4 \\
        Youth & 4 \\
        Death & 3 \\
        Equality & 3 \\
        Fate and chance & 3 \\
        Justice & 3 \\
        Marriage & 3 \\
        Mistakes & 3 \\
        Money, wealth, and prosperity & 3 \\
        Revenge and retribution & 3 \\
        Strength & 3 \\
        Violence & 3 \\
        Ambition & 2 \\
        Boastfulness and conceit & 2 \\
        Class & 2 \\
        Embarrassment, shame, and humiliation & 2 \\
        Food & 2 \\
        Gossip and rumor & 2 \\
        Madness & 2 \\
        Traitors and treachery & 2 \\
        Warfare & 2 \\
        Weakness & 2 \\
        Weather & 2 \\
        Jealousy and envy & 1 \\
        Pregnancy & 1 \\
        Clothes & 0 \\
        Debt & 0 \\
        Hypocrisy & 0 \\ 
    \bottomrule
    \end{tabular}
    \caption{Cont': EPIE-ME Idiomatic Expression Count ($ < 10$) per Theme}
    \label{tab:all_themes_2}
\end{table}

\section{Example Prompts}\label{app:prompts}
We provide all prompts used for our LLM experiments in Tab.~\ref{app:prompts}. The prefixes were used when exploring configurations. The contextual enhancement prompts are shown in zero-shot format, but were also used in the $k$-shot scenario. We assume the reader can infer the $k$-shot form based on the other $k$-shot templates.

\begin{table}[!ht]
\small
     \resizebox{\columnwidth}{!}{\begin{tabular}{>{\arraybackslash}m{2cm}|>{\arraybackslash}m{13cm}}
     %\begin{tabular}[ll]
     \toprule
       \textbf{Prompt Type}  &  \textbf{Text} \\
       \midrule
       Length control prefix &  
       \begin{itemize}[leftmargin=*,nosep]
           \item Single sentence: \texttt{Provide responses no longer than one sentence.}
           \item Two sentences max: \texttt{Provide responses no longer than two sentences.}
       \end{itemize}
       \\
       \midrule
       Zero-Shot & \begin{itemize}[leftmargin=*,nosep]
           \item Direct: \texttt{Explain in \{lang\} the meaning of the idiomatic expression \lq\{idiom\}'.}
           \item Pipeline-Explain: \texttt{Explain the meaning of the idiomatic expression \lq\{idiom\}'.}
           \item Pipeline-Translate: \texttt{Translate the following into \{lang\}: \lq\{explanation\}'. Respond with nothing but the translation.}
       \end{itemize}
       \\
       \midrule
       Contextual Enhancements & \begin{itemize}[leftmargin=*,nosep]
           \item Categorical information: \texttt{Explain in \{lang\} the meaning of the idiomatic expression \lq\{idiom\}' in the context of \lq\{category\}'.}
           \item Sentence-level context: \texttt{Explain the meaning of the idiomatic expression \lq\{idiom\}' given the sentence \lq\{sentence\}'.}
           \item Both: \texttt{Explain in \{lang\} the meaning of the idiomatic expression \lq\{idiom\}' in the context of \lq\{category\}' and given the sentence \lq\{sentence\}'.}
       \end{itemize}
       \\
       \midrule
       $k$-Shot & 
       All are repeated $k$ times with a $k+1$th occurrence having the idiom to be explained followed by \texttt{A: }
       \begin{itemize}[leftmargin=*,nosep]
           \item Direct: \texttt{Q: Explain in \{lang\} the meaning of the idiomatic expression \lq\{idiom\}'. A: \{explanation\}.}
           \item Pipeline-Explain: \texttt{Q: Explain the meaning of the idiomatic expression \lq\{idiom\}'. A: \{explanation\}.}
           \item Pipeline-Translate: \texttt{Q: Translate the following into \{lang\}: \lq\{explanation\}'. A: \{explanation\}.}
       \end{itemize}
       \\
       \midrule
       Fill-in-the-Blank &
       \begin{itemize}[leftmargin=*,nosep]
            \item System: Fill in the appropriate idiomatic expression that completes the sentence below.
            \item User: \{context\} Response: [blank] can be replaced with:
       \end{itemize}
       \\
       \midrule
       Oxford-ME Context Generation &
       \begin{itemize}[leftmargin=*,nosep]
            \item Use the idiom \{idiom\} meaning \{explanation\} in a sentence. Respond with nothing but the sentence.
       \end{itemize}
       \\
       \bottomrule
    \end{tabular}}
    \caption{LLM Prompts}\label{tab:prompts}
\end{table}

\begin{figure*}[!ht]
    \centering
    % pdf for arxiv, svg for ACL
    % \includesvg[width=0.7\textwidth]{noise.svg}
    \includegraphics[width=0.7\textwidth]{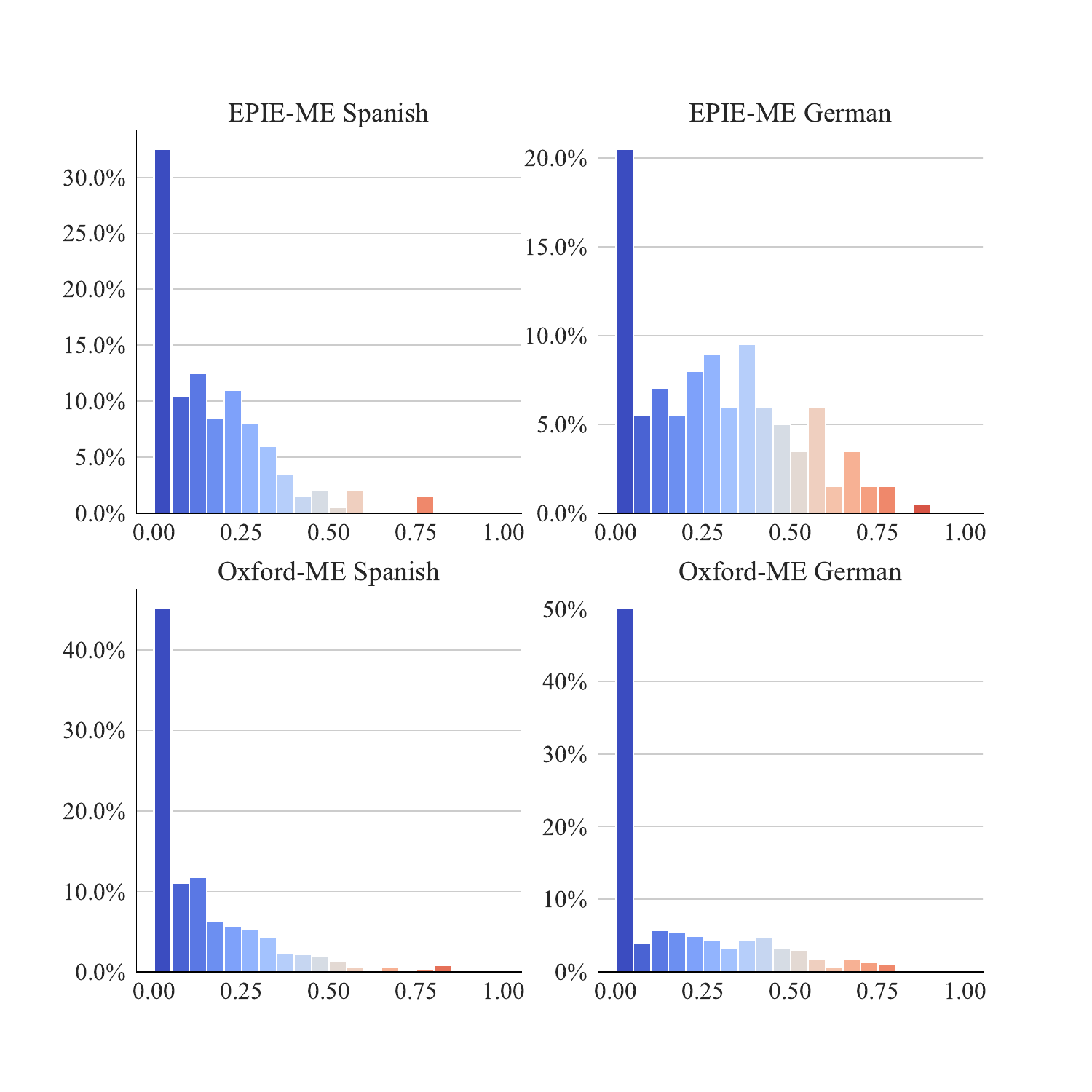}
    \caption{Histograms of Normalized Edit Distance vs Percentage of Data with Bins of Size 0.05 Over Both Languages in EPIE-ME and Oxford-ME.}
    \label{fig:noise}
\end{figure*}

\begin{table}[!t]
\centering
\small
\begin{tabular}{c|cc|cc}
\toprule
\textbf{Statistic} & \multicolumn{2}{c|}{\textbf{EPIE-ME}} & \multicolumn{2}{c}{\textbf{Oxford-ME}}\\
& Spanish & German & Spanish & German \\
\midrule
Mean & 0.159 & 0.283 & 0.131 & 0.183 \\
Min & 0 & 0.092 & 0 & 0 \\
Max & 0.781 & 0.857 & 0.8 & 0.889 \\
Stdev & 0.162 & 0.216 & 0.165 & 0.241 \\
\toprule
Q1 & 0 & 0.092 & 0 & 0 \\
Q2 & 0.131 & 0.269 & 0.071 & 0.043 \\
Q3 & 0.249 & 0.440 & 0.205 & 0.307 \\
Q4 & 0.781 & 0.857 & 0.839 & .89 \\
\bottomrule
\end{tabular}
\caption{Noise in the Data Measured by Normalized Edit Distance with Noise by Quartile in the Lower Portion.}\label{tab:noise}
\end{table}

\section{Noise in Translations}\label{app:noise}

Where EPIE-ME needed a small amount of correction, Oxford-ME required significantly less modification.\footnote{For Oxford-ME, we verify only the Spanish translations for this analysis.} We believe this is primarily due to the difference in English explanation length between the two sources of data. Examining the quartiles of normalized edit distance, we see that 50\% of the data required roughly less than a 14\% modification to the Spanish explanations in EPIE-ME, less than a 27\% modification for German in EPIE-ME, less than an 8\% modification for Spanish explanations in Oxford-ME, and less than a 4\% modification for German in Oxford-ME. Overall, quite low. The high standard deviations can be explained by the lack of samples near the mean, rather a large portion of samples are in the lower quartiles. These measurements can be found in Tab.~\ref{tab:noise}. Graphs of the same information can be found in Fig.~\ref{fig:noise}.

\section{Translation as a Method}\label{app:translation}

To test our assumption that translation is not a wholesale solution, we ask GPT-3.5 Turbo to provide only a translation of the idiomatic expressions in our test set. When comparing the semantic similarity of the generated translation to the gold explanation, we see that this approach appears to perform better than that of the fine-tuned models used in our experiments in Tab~\ref{tab:main_results}, earning scores of 45.80 and 44.21 for Spanish and German respectively. However, we asked our annotators to take a closer look at the outputs to make note of 3 things for each sample: \textit{Is the translated expression good/natural? Is the translation done word-for-word? Is this translation an instance of an idiom in your native language?}

Using these 3 factors, we analyze how often a sample is above a threshold of 40\footnote{We found the scores had incredibly high standard deviations of 20.52 and 19.84 respectively, so we set this threshold at a rough 10\% under the mean.} while still being judged as a poor/unnatural expression, which was the case for 14.5\% and 18.5\% of our test set in Spanish and German respectively. While semantic similarity is not meant to represent the actual quality of a sample, this further indicates that the similarity score we obtain cannot be the only metric used for evaluating models on this task. Of the total number of expressions that were good/natural, the proportion that were word-to-word translations were 48.3\% and 56.3\%, where 71.4\% and 93.1\% of those subsets were a case where the idiom was also present in the target language. This tells us that translation of just an idiomatic expression is not likely to generalize, as one will often need the same idiom to be present in the target language. These findings also hint at the potential that models trained to capture multilingual sentence similarity in their embedding spaces are more strongly capturing word-to-word similarity than meaning similarity in the case of idiomatic expressions.

\begin{table}[!t]
    \centering
    \small
    \begin{tabular}{l|cc}
        \toprule
        \multicolumn{3}{c}{\textbf{Correlation with Similarity Scores}} \\
        \toprule
        Model / Proxy & $\rho$ & $\tau$-b \\
        \midrule
        GPT vs SR & -0.181 & -0.123 \\
        Llama vs SR & -0.293 & -0.205 \\
        Llama vs GP & -0.030 & -0.021 \\
        \toprule
        \multicolumn{3}{c}{\textbf{Correlation with Human Evaluation}} \\
        \toprule
        Metric / Proxy & \multicolumn{2}{c}{$\tau$-b} \\
        \toprule
        Fluency vs SR & \multicolumn{2}{c}{-0.141} \\
        Accuracy vs SR & \multicolumn{2}{c}{-0.067} \\
        Fluency vs GP & \multicolumn{2}{c}{-0.159} \\
        Accuracy vs GP & \multicolumn{2}{c}{0.072} \\
        \bottomrule
    \end{tabular}
    \caption{EPIE-ME Model Similarity vs Search Results (SR) and Generation Probabilities (GP)}
    \label{tab:contamination}
\end{table}

\section{Data Contamination}\label{app:contamination}

Some may be skeptical of our experiments and the potential for data contamination in our LLMs. Specifically, that they may have seen the idiom-explanation pairs within our dataset. While we do not have access to the training data of either GPT or Llama, we will assume that they have seen almost the entire internet for the purposes of determining the level of contamination. Under this assumption, if the model is performing well on certain idiomatic expressions, we would expect those to be the most used across the internet. For both GPT and Llama, we attempt to correlate both the similarity score and human evaluation results per expression with the number of search results returned by the Google API when using the expression as input -- not accounting for morphological shift or interjecting words. For Llama, we additionally extract the probability of generating an expression given 10 in-context examples via fill-in-the-blank prompting.\footnote{Prompts found in Tab.~\ref{tab:prompts}} Our similarity score results are correlated with the search results and generation probabilities using Spearman's rank correlation coefficient ($\rho$) and Kendall's rank correlation coefficient, variant b ($\tau$-b). Because of the many ties in our human evaluation results, we correlate only with $\tau$-b. In every single case, we see a score very close to 0, indicating no association whatsoever (Tab.~\ref{tab:contamination}). We further examine our results by quartile of search result hits. There is no clear upward trend as search results increase. In fact, the Llama results show an inverse relationship, the Fluency results decrease until the final quartile, and the GPT and Accuracy results show a bell curve peaking in the second quartile (Tab.~\ref{tab:contam_q}).

\begin{table}[H]
    \small
    \centering
    \begin{tabular}{l|cccc}
        \toprule
        \textbf{Metric} & \textbf{Q1} & \textbf{Q2} & \textbf{Q3} & \textbf{Q4} \\
        \midrule
        GPT Similarity & 71.76 & 72.60 & 69.71 & 67.62 \\
        Llama Similarity & 63.40 & 61.99 & 60.85 & 55.81 \\
        Fluency & 4.85 & 4.70 & 4.58 & 4.67 \\
        Accuracy & 4.78 & 4.86 & 4.80 & 4.69 \\
        \bottomrule
    \end{tabular}
    \caption{EPIE-ME Results by Quartile of Search Results}
    \label{tab:contam_q}
\end{table}

%TODO or TOTAKEOUT as needed

\end{document}